\pdfoutput=1

\documentclass[11pt]{article}

\usepackage{ACL2023}

\usepackage{times}
\usepackage{latexsym}
\usepackage{graphicx}
\usepackage{multirow}

\usepackage{titlesec} 
\titleformat{\subsubsection}[runin]
  {\normalfont\normalsize\bfseries}{\thesubsubsection}{1em}{}

\usepackage[T1]{fontenc}

\usepackage[utf8]{inputenc}

\usepackage{microtype}

\usepackage{inconsolata}

\title{Real-Time Textless Dialogue Generation}

\author{Long Mai and Julie Carson-Berndsen\\
  ML-Labs, School of Computer Science, University College Dublin, Ireland \\
  \texttt{long.mai@ucdconnect.ie, julie.berndsen@ucd.ie} \\}

\begin{document}
\maketitle
\begin{abstract}
Recent advancements in large language models (LLMs) have led to significant progress in text-based dialogue systems. These systems can now generate high-quality responses that are accurate and coherent across a wide range of topics and tasks. However, spoken dialogue systems still lag behind in terms of naturalness. They tend to produce robotic interactions, with issues such as slow response times, overly generic or cautious replies, and a lack of natural rhythm and fluid turn-taking. This shortcoming is largely due to the over-reliance on the traditional cascaded design, which involve separate, sequential components, as well as the use of text as an intermediate representation. This paper propose a real-time, textless spoken dialogue generation model (RTTL-DG) that aims to overcome these challenges. Our system enables fluid turn-taking and generates responses with minimal delay by processing streaming spoken conversation directly. Additionally, our model incorporates backchannels, filters, laughter, and other paralinguistic signals, which are often absent in cascaded dialogue systems, to create more natural and human-like interactions. The implementations and generated samples are available in our repository\footnote{https://github.com/mailong25/rts2s-dg}.

\end{abstract}

\section{Introduction}

The success of large general-purpose language models, such as ChatGPT and LLAMA, has led to significant advancements in various natural language processing (NLP) tasks, especially in challenging tasks such as open-domain dialogue generation. Chatbot applications like Replika \cite{replika} and Character AI \cite{character} have demonstrated substantial progress in this sub-task, often achieving or surpassing human-level performance in terms of semantic coherence. Despite these advancements, conversations with current chatbots often feel unnatural, especially in spoken interactions. This issue primarily arises from the design of most chatbots, which prioritize generating well-written, formal responses, without incorporating informal conversational elements such as backchannels, laughter, and other paralinguistic cues. These elements are crucial for creating a more engaging, human-like experience.

\begin{figure*}
    \centering
    \includegraphics[width=\linewidth]{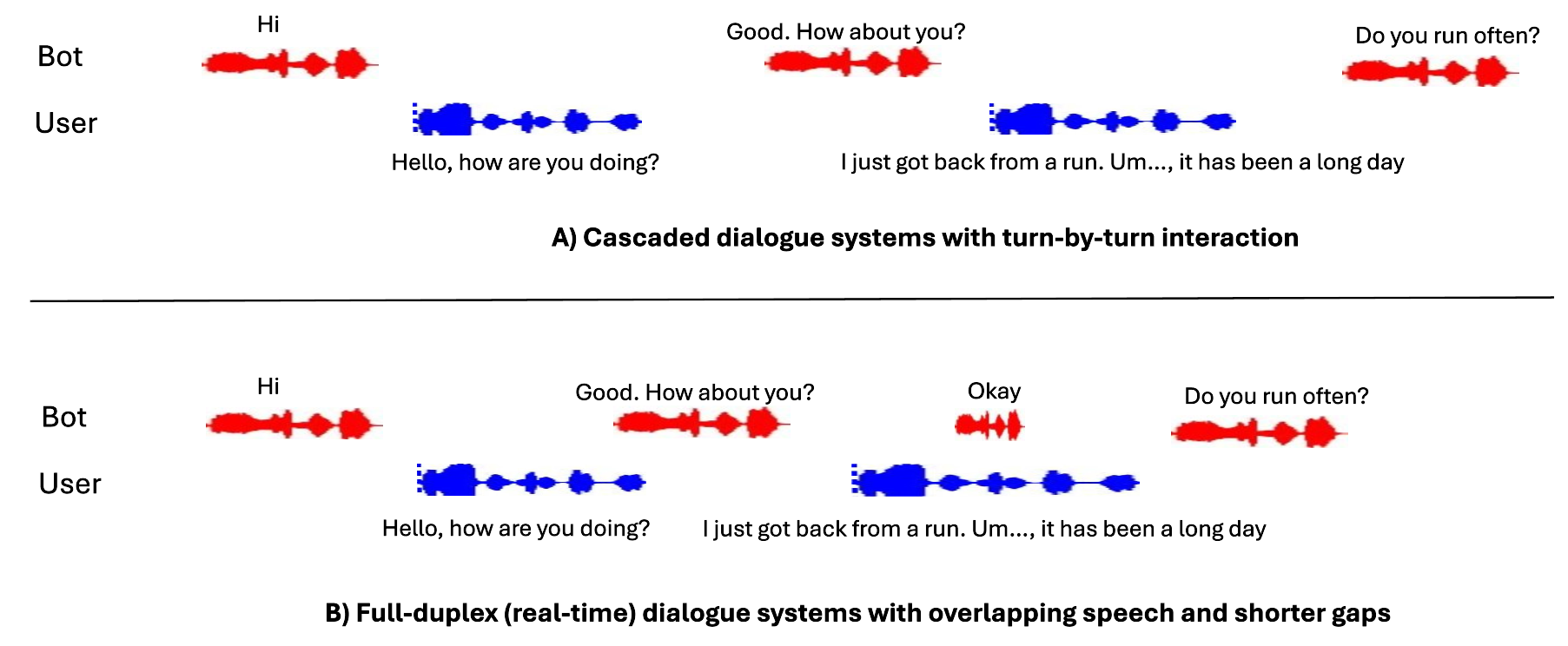}
    \caption{Examples of conversations with cascaded and real-time dialogue systems}
    \label{fig:conv_exam}
\end{figure*}

Most spoken-based systems rely on a cascade design, consisting of three distinct components: Automatic Speech Recognition (ASR), Dialogue Response Generation (DRG), and Text-to-Speech (TTS). This sequential structure introduces challenges for providing a fluid, interactive conversation, as it assumes turn-by-turn dialogue with no overlapping speech. In real-life conversations, however, two people can speak simultaneously through parallel channels. Figure \ref{fig:conv_exam} shows an example of conversation with cascaded dialogue system. In this system, the dialogue context is often expanded sequentially where each speaker’s utterance is concated one after another (e.g., Speaker1-Speaker2-Speaker1). This forces the response generation model to wait for the user's utterance to finish before it can generate a response, often with a delay (e.g., 800ms). This results in slow and unnatural conversational flow. Additionally, the reliance on text as an intermediate representation introduces increased latency and complexity when integrating multiple components. Generating spoken responses based on text representations also limits the expressiveness of the speech.

There is existing research on making spoken dialogue more naturalistic, including turn-taking prediction \cite{ekstedt2022much, gaze_cues}, backchannel generation \cite{hara_bc_multi, ishii_bc_multi}, and duplex conversations \cite{alibaba_duplex}. However, these approaches primarily use text as their input/output modality and focus more on task-oriented rather than open-domain generation. The work most similar to ours is dGSLM \cite{dgslm}, in which the authors propose a "textless" model capable of generating naturalistic spoken dialogue given a conversational prompt between two speakers. However, this model can only generate dialogue offline, making it impractical for real-world applications that require generating instant responses from streaming conversational input.

To overcome these limitations, we propose a novel approach: a real-time, textless dialogue generation model (RTTL-DG). Unlike traditional cascade systems, which require speech-to-text conversion before generating responses, our approach integrates ASR and DRG into a unified model that directly encodes spoken conversations. The encoded information is then used to predict the chatbot's next action—such as remaining silent or speaking—at regular, short intervals (e.g., 160ms). This enables fluid turn-taking and faster response times. Moreover, our model generates speech units instead of text as the target output, resulting in more natural-sounding speech. In summary, the RTTL-DG model provides the following advantages:

\begin{enumerate}
    \item \textbf{Real-time response generation.} The RTTL-DG model provides near-instantaneous responses by directly processing streaming input from conversations with minimal delay. 
    \item \textbf{Textless modeling.} Our approach eliminates the need for text as an intermediate representation, reducing latency introduced by the ASR system and simplifying the inference pipeline.
    \item \textbf{Spontaneous expressions.} The model generates natural, coherent responses, including informal conversational elements such as hesitations and laughters, which are essential for fostering human-like interactions.
    \item \textbf{Fluid turn-taking.} RTTL-DG enables smooth turn-taking in conversations, capable of handling interruptions and speech overlaps.
    \item \textbf{Diverse response formats.} The model produces responses of varying lengths, from short back-and-forth phrases to long-form storytelling formats.
\end{enumerate}

Experimental results show that the proposed RTTL-DG model provides a more interactive and natural conversational experience. While it may slightly underperform in terms of semantic coherence compared to cascade models, it excels in naturalness, responsiveness, and fluidity. We believe these advancements are crucial for developing more engaging and relatable chatbot systems. The RTTL-DG model holds significant potential across various domains, including companionship, therapy, customer service, and other areas that require human-like dialogue interactions.

\section{RTTL-DG inference pipeline}

\begin{figure*}
    \centering
    \includegraphics[width=14cm]{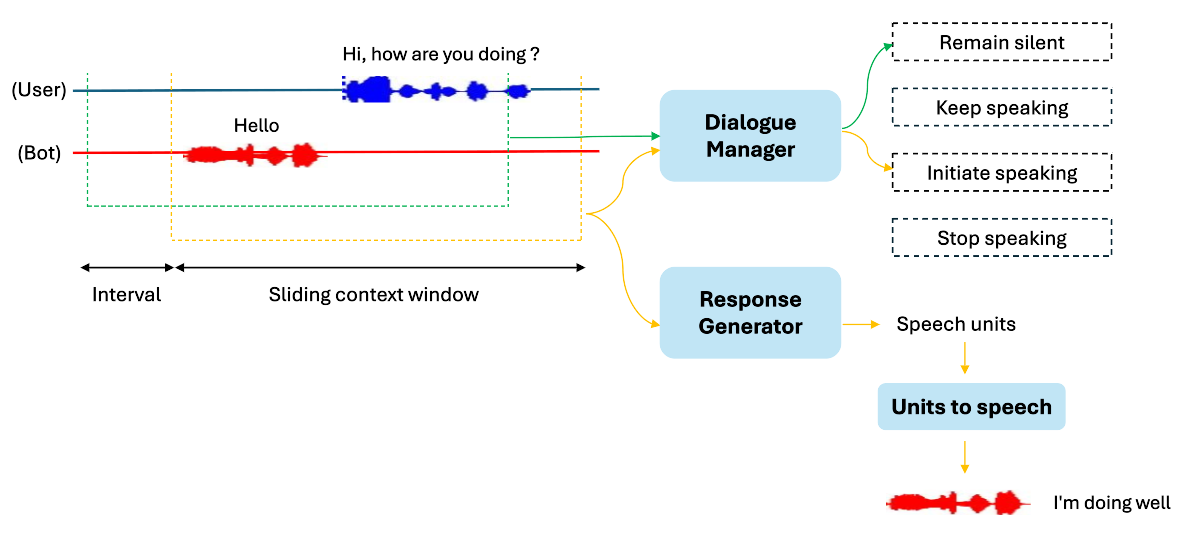}
    \caption{Inference pipeline of RTTL-DG}
    \label{fig:s2s_inference}
\end{figure*}

Figure \ref{fig:s2s_inference} illustrates the process by which RTTL-DG generates responses during inference. The system comprises two main components: the Dialogue Manager and the Response Generator. Both modules process continuous streaming speech inputs from both the user and the chatbot itself. 

At regular intervals (i.e., every 160 ms), the Dialogue Manager determines the chatbot's next action based on the most recent segment of the streaming input, truncated by a sliding context window (e.g., the last 20 seconds). The Dialogue Manager must select one of four possible actions: (1) Remain silent: the chatbot continues to listen without responding, (2) Initiate speaking: the chatbot begins to generate a response, with the content provided by the Response Generator, (3) Keep speaking: if the chatbot is already speaking, this action directs it to continue its current utterance, (4) Stop speaking: if both the user and the chatbot are speaking simultaneously, this action halts the chatbot’s speech. Actions (1) and (2) enable the chatbot to determine the optimal moment to respond, which does not necessarily rely on detecting the end of the user’s turn (e.g., a fixed period of silence). This flexibility allows the chatbot to provide timely feedback, such as backchannels, even while the user is still speaking. Meanwhile, actions (3) and (4) ensure smooth handling of interruptions, enabling the chatbot to adapt dynamically to user input without compromising the flow of conversation.

When the Dialogue Manager decides to initiate a response, the Response Generator takes over. It uses the same input features as the Dialogue Manager-the most recent segment of the streaming conversation from both speakers. Instead of generating traditional text-based responses, this model generates speech units. The discovery of these units is done by training a speech quantization model (i.e., HUBERT+KMeans \cite{hubert}) on a large amount of unlabeled speech audio. Using speech units allows the chatbot to generate more natural and dynamic expressions, including elements such as pauses, hesitations, backchannels, and even laughter. Once the Response Generator generates the speech units, a unit-to-speech model converts them into a final spoken response.

\section{RTTL-DG architecture}
\label{sec:s2s_design}

\begin{figure*}
    \centering
    \includegraphics[width=15cm]{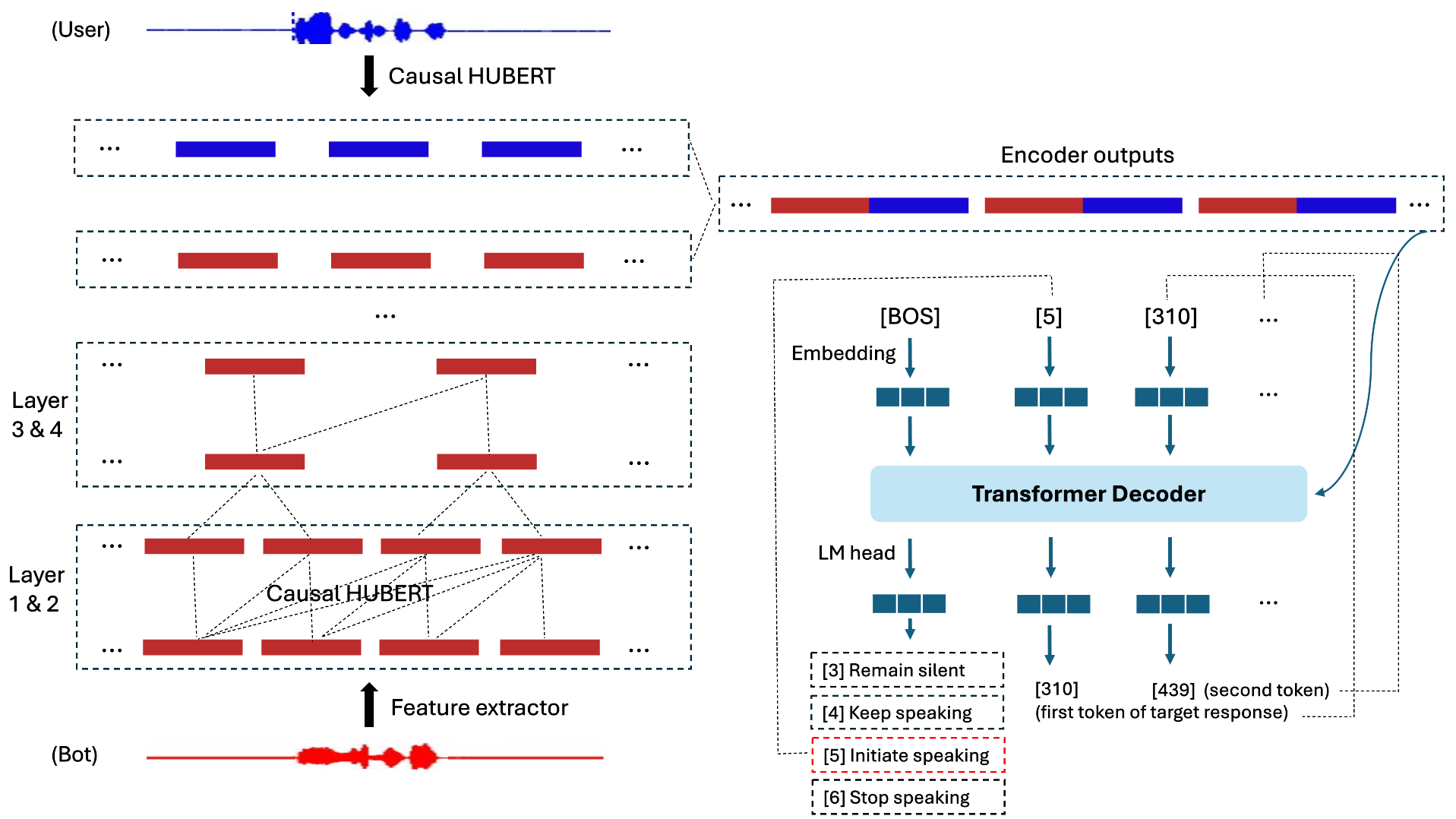}
    \caption{Model architecture of RTTL-DG}
    \label{fig:s2s_model}
\end{figure*}

Figure \ref{fig:s2s_model} illustrates the model architecture of RTTL-DG, which comprises two main components: an encoder and a decoder.

The encoder is designed as a streaming causal transformer, tasked with processing the continuous speech input from each speaker. It includes a convolutional feature extractor followed by eight transformer encoder layers. The convolutional feature extractor converts raw audio signals into a sequence of hidden states with dimensions $R^{T \times D}$, where $T$ represents the number of timesteps (or sequence length), and $D$ is the hidden dimension, set to 768. Each timestep corresponds to a 20ms segment of audio. The transformer encoder layers further refine these hidden states into contextual representations. To improve efficiency, the number of timesteps is progressively reduced by merging two adjacent hidden states after layers 2, 4, and 6. This reduction not only speeds up computation but also enhances the semantic richness of the hidden states. Since the model processes streaming input, causal masking is applied to ensure that the encoder layers do not access future information. The final output of the encoder for each speaker has dimensions $R^{(T/8) \times D}$. Positional embeddings and speaker embeddings, both with dimensions $R^{(T/8) \times D}$, are then added to the encoder outputs to incorporate temporal and speaker-specific information. As the outputs from each speaker are time-aligned, they can be concatenated to produce the final encoder output $R^{(T/8) \times 2D}$. Each timestep in this final representation corresponds to a 160ms chunk of speech, containing encoded information from both speakers. We initialize the encoder using a pre-trained HUBERT model \cite{dgslm}, which was trained on 2000 hours of the Fisher dataset \cite{fisher}.

The decoder plays a dual role as the Dialogue Manager and the Response Generator. It takes the streaming input \(R^{(T/8) \times 2D}\) and, at every 160ms interval, determines the next action of the chatbot, while also generating the content of its response if required. The decoder architecture consists of an embedding layer, eight transformer decoder layers, and a projection language model head. The embedding layer converts input token indices into embeddings with dimensions \(R^{S \times 2D}\), where \(S\) is the target sequence length and $2D$ is the hidden dimension. These embeddings are passed through the decoder layers, which extract contextual information and attend to the encoder outputs \(R^{(T/8) \times 2D}\) from both speakers. The final contextualized decoder hidden states, with dimensions \(R^{S \times 2D}\), are projected into \(R^{S \times V}\) by the language model head, where \(V\) is the vocabulary size. The vocabulary includes special tokens for representing the chatbot's next actions: padding (0), beginning of response [BOS] (1), end of response [EOS] (2), remain silent (3), continue speaking (4), start speaking (5), and stop speaking (6). The remaining token indices in the vocabulary represent the content of the chatbot's verbal response. More specifically, when the Dialogue Manager decides that the chatbot should initiate a speech, the first token generated by the decoder should be 5, followed by tokens that form the chatbot’s verbal response.

\section{Training data for RTTL-DG}
\label{sec:s2s_data}

\begin{figure*}
    \centering
    \includegraphics[width=11cm, height=8cm]{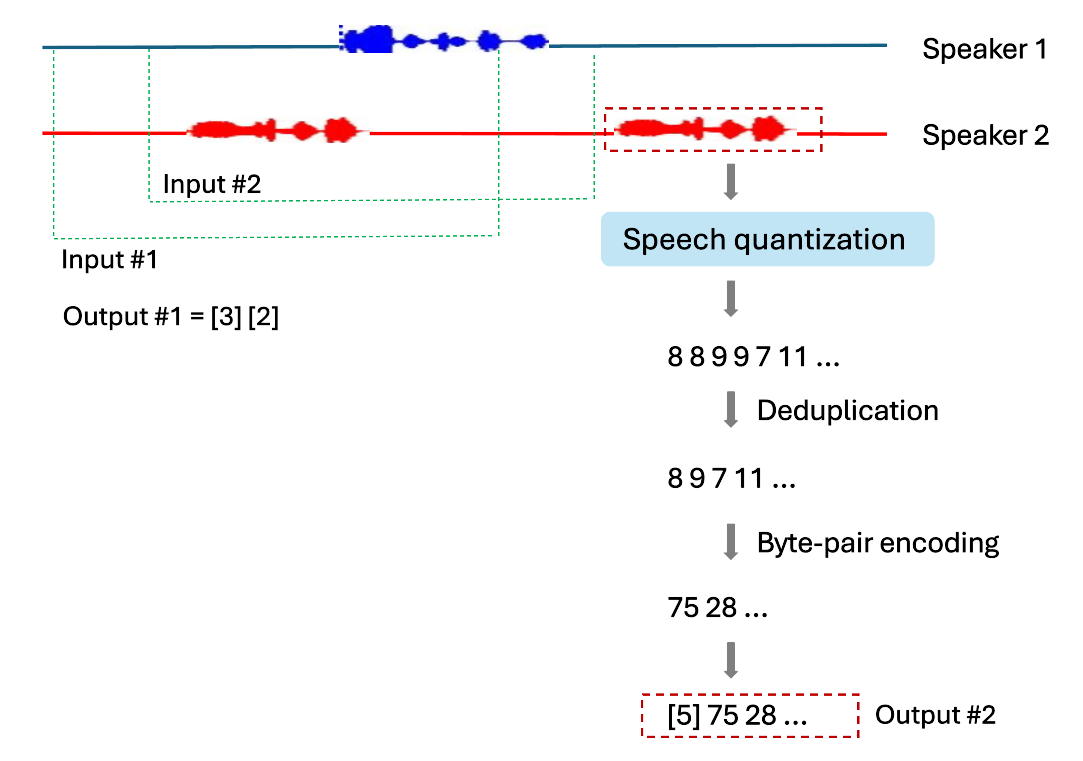}
    \caption{An example of how a training sample is constructed for RTTL-DG}
    \label{fig:s2s_data}
\end{figure*}

Figure \ref{fig:s2s_data} illustrates example pairs of training samples used for training the RTTL-DG model. As mentioned earlier, for each interval (i.e., 160 ms), there is a training pair of (input, output). The input is the preceding truncated conversation from both speakers, which is then encoded using the encoder introduced in Section \ref{sec:s2s_design}. The output is a sequence of tokens containing two types of information: the next action and the response content.

The first token in the output represents the next action (e.g., 3 indicates remaining silent, and 5 indicates initiating a response). To assign the next action label for each input, we rely on its last segment (\( s_{last} \)), which is the last 160ms chunk of the input. For each speaker, we first use voice activity detection (VAD) \cite{funasr} to identify speech segments (\( SE \)) and silent segments (\( SI \)). If \( s_{last} \) falls entirely within a silent segment, the next action is labeled as remaining silent. If \( s_{last} \) lies completely within a speech segment, the action is labeled as continuing to speak. When \( s_{last} \) marks the beginning of a speech segment, the action is labeled as initiating speech. If \( s_{last} \) occurs at the end of a speech segment and there is overlapping speech between the two speakers, the action is labeled as stop speaking.

If a response is initiated, the subsequent tokens in the output represent the response content. Instead of using text tokens for the target response, we use speech units extracted from a pre-trained speech quantization model (i.e., HUBERT+K-means) \cite{dgslm}. This approach enables the generation of more natural and expressive speech, including representations of non-verbal units such as laughter, pauses, and hesitations. To reduce the target response length and increase the semantic richness of each token, we apply deduplication (collapsing consecutive identical tokens into one) followed by byte-pair encoding to merge adjacent tokens into larger units. This significant reduction in output length facilitates faster response generation, thereby lowering overall response latency.

We use the Switchboard-2 Phase II corpus \cite{switchboard} to generate the training dataset. This dataset consists of 4,472 five-minute telephone conversations between two speakers discussing various topics, such as jobs, hobbies, and movies. As shown in Table \ref{tab:s2s_data}, the amount of data in Switchboard is relatively limited—approximately 372 hours. This small dataset presents significant challenges for modeling open-domain response generation, which typically requires millions of training instances \cite{blender}. To address this problem, we propose a two-stage approach: first, pretraining the RTTL-DG model on a synthetic dataset to learn the response generation task, and then fine-tuning it on Switchboard data to adapt to the next action prediction task.

To create the synthetic dataset, we leverage the abundance of text-based two-party conversations. Initially, we extract 91,796 five-turn conversations from several datasets, including BlendedSkillTask \cite{blender}, ConvAI \cite{convai}, TopicalChat \cite{topical}, EmpatheticDialogues \cite{empathetic}, and WizardOfWikipedia \cite{wow}. We then use GPT-4o to extend each conversation by generating an additional 35 turns. Finally, a text-to-speech model is used to convert these extended text-based conversations into speech-based conversations, resulting in 5,798 hours of synthetic data, with an average duration of four minutes per conversation. The text-to-speech model employed is a multi-speaker VITS2 model \cite{vits2}, trained using Switchboard data to ensure minimal differences between the synthetic and real datasets.

\begin{table*}[]
\centering
\begin{tabular}{ccccccccc}
\hline
\multirow{2}{*}{Dataset} & \multirow{2}{*}{Hours} & \multirow{2}{*}{\begin{tabular}[c]{@{}c@{}}Number of\\ dialogues\end{tabular}} & \multirow{2}{*}{\begin{tabular}[c]{@{}c@{}}Number of \\ speakers\end{tabular}} & \multicolumn{4}{c}{\begin{tabular}[c]{@{}c@{}}Training instances\\ per conversation\end{tabular}} & \multirow{2}{*}{\begin{tabular}[c]{@{}c@{}}Response \\ length\end{tabular}} \\ \cline{5-8}
                         &                        &                                                                                    &                                                                                & SIL                     & CON                     & SPK                   & STP                   &                                                                             \\ \hline
Switchboard              & 372                    & 4472                                                                               & 679                                                                            & 1751                    & 1395                    & 112                   & 70                    & 2.8s                                                                        \\ \hline
Synthetic                & 5798                   & 91796                                                                              & 300                                                                            & 1509                    & 1287                    & 41                    & 0                     & 4.9s                                                                        \\ \hline
\end{tabular}
\caption{Summary of the characteristics of synthetic and real datasets. SIL, SPK, CON, and STP denote Remain Silent, Initiate Speaking, Keep Speaking, and Stop Speaking, respectively.}
\label{tab:s2s_data}
\end{table*}

\section{Experiment settings}
This section describes the setups we use to build and evaluate the proposed real-time dialogue generation models. We also describe the baseline cascaded model used for comparison.
\subsection{Evaluation metrics}

There are two settings we use to evaluate the dialogue generation model: single-turn and multi-turn settings. In the single-turn setting, we provide the model with a spoken dialogue context and ask it to predict the next action and response. In the multi-turn one, we provide the model with a seeded conversation between two speakers and ask it to complete the conversation for the next 30 seconds.

\ 

\noindent \textbf{Single-turn metrics.} As mentioned earlier, at regular intervals (i.e., every 160 ms), the model determines the chatbot's next action and generates response content based on the most recent streaming input. For the next action recognition task, we evaluate the model using Precision, Recall, and F1-score metrics. These metrics assess how effectively the model predicts one of four possible actions: remain silent (SIL), initiate speaking (SPK), keep speaking (CONT), or stop speaking (STP). The predictions are compared against ground-truth labels derived from the Switchboard test set, providing a benchmark for measuring the model's performance in understanding and managing conversational flow. For the response generation task, we evaluate the model's performance in terms of semantic coherence and naturalness. Semantic coherence measures how logical and relevant the generated responses are within the dialogue context. To assess coherence, we provide GPT-4o with a dialogue context and a generated response, then prompt it to assign a coherence score on a scale from 1 (worst) to 10 (best). We report the average coherence score, along with the percentage of responses rated as sensible—those with a coherence score of 5 or higher. We use Whisper \cite{whisper} to transcribe the spoken context and response into text before prompting GPT-4o for coherence assessment.

To evaluate the naturalness of the spoken responses, we use the following metrics:

\begin{enumerate}
    \item \textbf{Pitch/Energy standard deviation (PSTD/ESTD).} Measures variation in pitch and energy levels. Higher values indicate more dynamic and natural-sounding speech with tonal variation.
    
    \item \textbf{Words per minute (WPM).} Tracks the pace of speech. A natural speaking speed varies by context, and speaking too fast or too slowly can sound unnatural.
    
    \item \textbf{Filler words per minute (FWPM).} Measures the frequency of filler words like "um" or "uh." While occasional fillers are normal in speech, excessive use can make the speaker sound hesitant.
    
    \item \textbf{Repetitions per minute (RPM).} Tracks repeated words or phrases. Some repetition is natural, but excessive repetition can make speech feel redundant or robotic.
    
    \item \textbf{Laughs/breaths per minute (LPM/BPM).} Measures the frequency of laughter and breaths in speech. Occasional laughter and breathing contribute to naturalness, but excessive amounts can seem forced.
    
    \item \textbf{Silence per minute (SPM).} Measures the average length of pauses. Short pauses are natural, but too many or too few can disrupt the speech flow.
\end{enumerate}

For each response generated by the cascaded and RTTL-DG models, we calculate the above metrics and average them across the entire test set. We then compute the same metrics for the ground truth (i.e. human-generated responses). A more effective model, in terms of naturalness, should generate statistics that more closely align with the ground truth. We use PRAAT software to calculate pitch and energy standard deviation. We prompt GPT-4o to count the number of filler words and repetitions in the transcribed text of each spoken response. For laughter and breathing events, we use a publicly pre-trained model based on the AudioSet dataset \cite{audioset}.

\ 

\noindent \textbf{Multi-turn metrics.} To assess chatbot performance across multiple conversational turns, we analyze interactions that mimic realistic dialogue scenarios. While deploying chatbots to engage with human participants offers valuable insights, this approach can be resource-intensive and time-consuming. To address this challenge, we employ a self-chat evaluation method. Starting with a seeded dialogue context, we proceed a conversation between two chatbots (based on the same model) that interact with each other for a set duration (i.e., 30 seconds). The resulting conversations are analyzed to evaluate how well the model performs in terms of naturalness, assessing its ability to approximate human-like dialogues. This method enables scalable testing and generates diverse conversational data without extensive human intervention.

\begin{figure*}
    \centering
    \includegraphics[width=14cm]{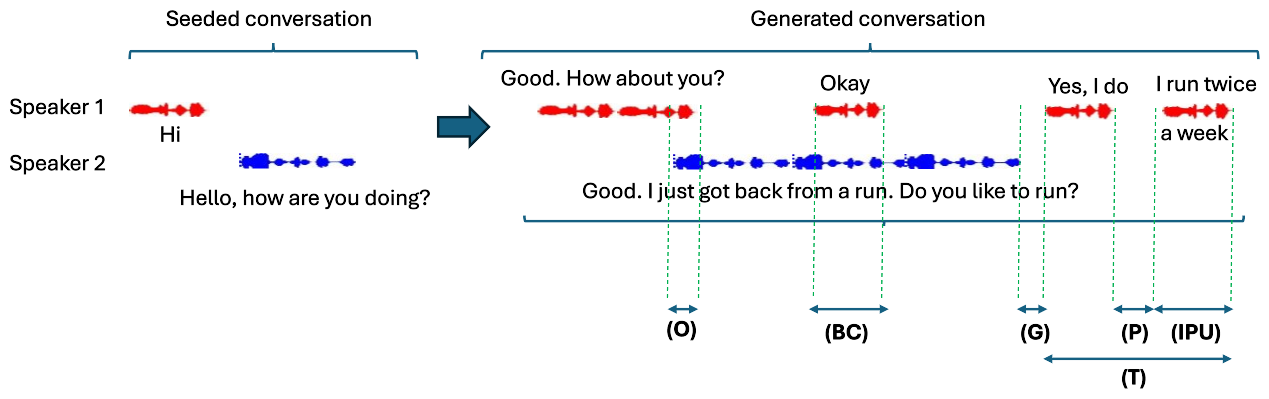}
    \caption{Multi-turn evaluation metrics: (IPU)-Interpausal Unit, (T)-Turn, (P)-Pause, (G)-Gap, (O)-Overlap, and (BC)-Backchannel.}
    \label{fig:eval_dialog}
\end{figure*}

Building on prior research \cite{dgslm}, we evaluate conversational naturalness using metrics such as overlap frequency, back-channels, pauses, and average gap durations. Figure \ref{fig:eval_dialog} illustrates examples of these metrics. We first apply VAD on the generated speech for each speaker. An Inter-Pausal Unit (IPU) is defined as a continuous segment of speech from one speaker. Successive IPUs separated by silences of less than 400 ms from the same speaker are grouped into a single turn. Within each turn, pauses are identified as silent segments lasting longer than 200 ms. Overlap refers to segments where both speakers are talking simultaneously. A specific type of overlap, called back-channeling, occurs when one speaker's short IPUs (less than 1 second) are entirely contained within the other speaker's turn, such as brief acknowledgments ("mm-hmm," "okay"). Gaps are defined as the silent intervals between turns from two different speakers. Gaps are also interpreted as latency, representing the time it takes for the next speaker to respond after the current speaker completes their turn. For overlaps, back-channels, and pauses, we count their occurrences per minute of conversation. For gaps, we calculate the average duration in milliseconds. These metrics provide a comprehensive framework to quantify the conversational dynamics and naturalness of chatbot interactions.

\subsection{Baselines}

We use a traditional cascaded dialogue generation model for comparison. This model consists of three main components: (1) an ASR system, which transcribes spoken dialogue context into text; (2) a text-based response generation model, which produces a next response based on the transcribed context; and (3) a TTS model, which converts the generated text responses back into speech. For the ASR system, we employ the state-of-the-art Whisper-Medium model \cite{whisper}. The text-to-speech component uses the multi-speaker VITS2 model, which is the same model described in Section \ref{sec:s2s_data}. The text-based response generation model is a transformer encoder-decoder, initially trained on synthetic conversational data and later fine-tuned on the Switchboard dataset. Details about the dataset are provided in Section \ref{sec:s2s_data}.

For single-turn evaluation, we assess the cascaded model's performance on response generation using semantic coherence and naturalness scores. In multi-turn evaluation, the cascaded model generates self-chat conversations sequentially. This means that the second speaker begins responding only after the first speaker has completed their turn. To detect the end of a speaker's turn, we apply a simple heuristic rule based on a fixed silence threshold of 800 ms, following the approach described in \cite{alibaba_duplex}. Specifically, if there is a silence of 800 ms after a speaker finishes, it is considered the end of their turn, allowing the second speaker to proceed with generating their response.

\subsection{Training configuration}

We used the Huggingface toolkit to train all the models. For the RTTL-RG model, the encoder was initialized with a pre-trained HuBERT model trained on the Fisher dataset \cite{dgslm}, while the decoder was randomly initialized. The encoder and decoder were configured with 8 layers each. The hidden dimensions were set to 768 for the encoder and 1536 for the decoder. The same hyperparameters were applied to the cascaded text-based response generation model.

Pretraining the RTTL-RG model on the synthetic dataset took approximately three days, and fine-tuning it on the Switchboard dataset required an additional day. Both processes were carried out using a single NVIDIA H100 GPU.

\section{Experiment results}

We present a detailed analysis of the performance and behavior of our proposed dialogue models through comprehensive experiments. The evaluation encompasses several key aspects, including next action prediction, semantic coherence, and naturalness of generated responses and dialogues. Examples of the generated outputs are available in our repository\footnote{https://mailong25.github.io/rts2s-dg}.

\begin{table*}[h]
\centering
\begin{tabular}{lccccc}
\hline
Next action       & Support & Precision & Recall & F1-score \\ \hline
Remain silent     & 1751    & 0.81      & 0.89   & 0.85     \\ \hline
Keep speaking     & 1539    & 0.94      & 0.96   & 0.95     \\ \hline
Initiate speaking & 112     & 0.62      & 0.43   & 0.52     \\ \hline
Stop speaking     & 42      & 0.75      & 0.53   & 0.62     \\ \hline
\end{tabular}
\caption{Next action prediction results for single-turn evaluation on Switchboard test set.}
\label{tab:next_action}
\end{table*}

\subsection{Next action prediction}

Table \ref{tab:next_action} shows the results for next action prediction on Switchboard test set. The Dialogue Manager demonstrates excellent performance in predicting the "Remain Silent" and "Keep Speaking" actions, with F1-scores of 0.85 and 0.95, respectively. These results reflect the system's proficiency in managing conversational continuity. The "Keep Speaking" action exhibits the highest F1-score, indicating the system excels at maintaining speech continuity when appropriate. The high precision (0.94) and recall (0.96) suggest the model effectively identifies when to continue speaking without unnecessary interruptions. This task is relatively straightforward, as the system often decides to keep speaking based on the presence of ongoing self-generated speech in recent intervals. Challenges primarily arise in scenarios with prolonged overlapping speech, y novel ideas are underrepresented the system must decide between continuing or halting. However, such scenarios are relatively rare in natural conversations, mitigating their overall impact on performance. Similarly, the model performs well in predicting when to remain silent, achieving a strong recall of 0.89, indicating that it frequently identifies opportunities to withhold responses appropriately. However, its slightly lower precision (0.81) suggests occasional over-prediction of silence, where the chatbot remains silent even when a response might be warranted. This behavior could occasionally hinder conversational engagement.

In contrast, the system struggles with predicting the "Initiate Speaking" and "Stop Speaking" actions, which are critical for maintaining adaptive and timely conversational responses. Predicting when to begin speaking is the most challenging task, with a precision of 0.62 and a recall of 0.43. When observing the confusion matrix, we found that a significant number of "Initiate Speaking" instances were misclassified as "Remain Silent," leading to missed opportunities for initiating responses. This issue can create awkward conversational silences or delays in the chatbot's contributions, negatively affecting conversational flow. Multiple factors contribute to this challenge. The limited number of training examples (only 112 instances per conversation) restricts the model’s ability to learn from these cases. Furthermore, open-domain conversations lack definitive rules for when to begin speaking or provide backchannels, and individual differences in speaking styles further increase the variability and noise in ground-truth labels.

\begin{table*}[]
\centering
\begin{tabular}{cccc}
\hline
Datasets                      & Models   & Coherence & Sensible responses \\ \hline
\multirow{2}{*}{Synthetic}   & Cascaded & 6.6            & 69\%                    \\ \cline{2-4} 
                             & RTTL-DG & 5.2             & 45\%                    \\ \hline
\multirow{2}{*}{Switchboard} & Cascaded & 6.4             & 69\%                    \\ \cline{2-4} 
                             & RTTL-DG & 4.8             & 48\%                    \\ \hline
\end{tabular}
\caption{Response generation result for a single-turn evaluation}
\label{tab:s2s_rg}
\end{table*}

\begin{table*}[t]
\centering
\begin{tabular}{llc}
\hline
\multicolumn{1}{c}{Models} & \multicolumn{1}{c}{Target response types} & Coherence \\ \hline
\multirow{2}{*}{Cascaded}  & Speech units                        & 5.6       \\ \cline{2-3} 
                           & Texts                               & 6.6       \\ \hline
\multirow{2}{*}{RTTL-DG}  & Speech units                        & 5.2       \\ \cline{2-3} 
                           & Texts                               & 6.2       \\ \hline
\end{tabular}
\caption{Performance of response generation with different target responses on a synthetic test set}
\label{tab:s2s_aba}
\end{table*}

Moderate performance is observed for the "Stop Speaking" action, with a precision of 0.75 and a recall of 0.53. While the system has some capacity to recognize when to halt speech, particularly in response to interruptions or overlapping dialogue, errors are prevalent in distinguishing "Stop Speaking" from "Keep Speaking." Such misclassifications highlight the challenges of managing interruptions effectively, which could lead to the chatbot either talking over the user or stopping unnecessarily, both of which disrupt conversational fluidity.

A major factor contributing to the system's challenges is the significant imbalance in class distribution. Actions like "Initiate Speaking" and "Stop Speaking" are underrepresented compared to the dominant "Remain Silent" and "Keep Speaking" classes. This imbalance limits the model’s ability to generalize effectively for the rarer actions, as evidenced by their lower performance metrics. Addressing this imbalance through techniques such as advanced data augmentation techniques, resampling, or the use of loss functions tailored for imbalanced datasets could improve model performance.

\subsection{Semantic evaluation results}

Table \ref{tab:s2s_rg} shows the semantic evaluation results for response generation task. As can be seen, both the Cascaded and RTTL-DG models achieve moderate coherence and sensible response scores, with room for improvement. The coherence scores for the Cascaded model are 6.6 and 6.4 for the synthetic and Switchboard datasets, respectively, while the RTTL-DG model achieves lower scores of 5.2 and 4.8 for the same datasets. These scores are still relatively modest, which likely reflects the limited size of the training data—both synthetic and real—which are far smaller in scale than the large datasets used for training recent state-of-the-art dialogue models. This suggests that larger, more diverse datasets could improve the performance of both models.

In terms of sensible responses, the Cascaded model consistently outperforms the RTTL-DG model, with scores of 69\% and 69\% on the synthetic and Switchboard datasets, respectively. The RTTL-DG model lags behind, with sensible response rates of 45\% and 48\% for the same datasets. The lower sensible response rates for RTTL-DG can be attributed to the complexity of the model, which generates speech units that include pauses, hesitations, and other dynamic speech features. This adds complexity to the generation process, which may impact coherence and the model’s ability to produce contextually appropriate responses.

Interestingly, both models perform better when trained and evaluated on synthetic data as opposed to real-world datasets like Switchboard. This is likely because synthetic data is cleaner, more structured, and better organized, allowing the models to more easily learn patterns and generate coherent responses. In contrast, real-world data is inherently noisier, more variable, and often more challenging for models to handle effectively, which can account for the lower performance on the Switchboard dataset.

To understand why the cascaded model outperforms the RTTL-DG model in terms of coherence, we conducted an ablation study on different types of target responses used to train the response generation model on the synthetic dataset, as shown in Table \ref{tab:s2s_aba}. When using the same type of target response, the cascaded model outperforms the RTTL-DG model for both text-based and unit-based responses. This suggests that using text as an intermediate representation enhances encoding performance, resulting in better coherence scores. However, it should be noted that this approach may increase latency, as an external streaming ASR model must be used as a frontend, often causing significant delays.

\begin{table*}[]
\centering
\begin{tabular}{ll}
\hline
\multicolumn{2}{l}{where are you}                                                                      \\
Units 1 & 45, 198, 117, 5, 61, 54, 44, 143, 196, 79, 37, 121, 40, 152, 97, 151, 175, 107, 88, 66, ..   \\
Units 2 & 45, 198, 117, 44, 85, 115, 196, 79, 37, 121, 40, 186, 152, 97, 151, 175, 107, 88, 66, ..     \\ \hline
\multicolumn{2}{l}{}                                                                                   \\ \hline
\multicolumn{2}{l}{yeah i am sure}                                                                      \\
Units 1 & 45, 198, 5, 87, 61, 81, 115, 196, 97, 100, 68, 142, 28, 17, 75, 3, 101, 34, 126, 56, 121, .. \\
Units 2 & 45, 198, 117, 44, 85, 115, 154, 196, 100, 68, 142, 28, 17, 75, 3, 101, 34, 126, 121, 40, ..  \\ \hline
\end{tabular}
\caption{Examples of speech unit representations from two different speakers when they utter the same sentence.}
\label{tab:unit_error}
\end{table*}

\begin{table*}[t]
\centering
\begin{tabular}{lcccccccc}
\hline
             & PSTD  & ESTD  & WPM & FWPM  & RPM  & LPM  & BPM  & SPM   \\ \hline
Cascaded     & 30.40 & 0.014 & 213 & 24.00 & 1.30 & 0.00 & 0.00 & 0.54s \\ \hline
RTTL-DG     & \textbf{32.20} & \textbf{0.019} & \textbf{172} & \textbf{23.20} & \textbf{6.40} & \textbf{0.50} & \textbf{0.22} & \textbf{1.66s}  \\ \hline
Ground Truth & 53.20 & 0.022 & 183 & 21.10 & 7.70 & 1.08 & 2.02 & 4.17s \\ \hline
\end{tabular}
\caption{Comparison between RTTL-DG and Cascaded in terms of naturalness on the Switchboard test set. PSTD and ESTD refer to the standard deviation of pitch and energy, respectively. WPM, FWPM, RPM, LPM, BPM, and SPM represent the average number of words, filler words, repetitions, laughs, and breaths per minute, respectively. SPM also refers to the average silence duration (in seconds) per minute.}
\label{tab:natural_single}
\end{table*}

When comparing different types of target responses, we observed that switching from text-based to unit-based responses significantly reduces coherence scores—from 6.6 to 5.6 in the cascaded system. Similar results were observed in the RTTL-DG model. This indicates that the use of unit-based responses is the primary reason RTTL-DG underperforms compared to the cascaded model, as previously shown in Table \ref{tab:s2s_rg}. This discrepancy can be attributed to the variability in speech unit representations. Factors such as speaker identity, rhythm, pitch, and energy introduce acoustic variability, making unit-based representations inconsistent. Evidence for this is provided when we compare the unit error rate between two speakers saying the same content across 156 sample pairs. The unit error rate is notably high at 50\%, indicating significant variability in the representations. This variability complicates modeling, especially when training data is limited. Table \ref{tab:unit_error} shows examples of variability in the speech units.

\subsection{Naturalness evaluation results}

Tables \ref{tab:natural_single} and Table \ref{tab:s2s_ipus} present the results of naturalness evaluations for dialogue generated by the cascaded model and the RTTL-DG model on real dataset. As shown in Table \ref{tab:natural_single}, RTTL-DG outperforms the Cascaded model across all naturalness metrics in the single-turn evaluation. Specifically, RTTL-DG generates more natural speech variations, with pitch and energy standard deviations closer to the ground truth. It also effectively replicates natural speech disfluencies such as repetitions, breathing, and laughter—features that the Cascaded model struggles to imitate. Additionally, RTTL-DG adjusts speech rate and pauses to better align with human speech, making its responses sound more natural and less robotic compared to the faster, more monotonous speech of the Cascaded model. These improvements are largely due to RTTL-DG's use of speech units as target responses, which enables more natural outputs.

\begin{table*}[t]
\centering
\begin{tabular}{lcccc}
\hline
Conversations & \multicolumn{3}{c}{Number of occurrence per minute} & Average gap \\ \hline
                                  & Overlaps        & Backchannels       & Pauses       &             \\ \hline
Cascaded                          & 0               & 0                  & 0            & 800ms       \\ \hline
RTTL-DG                          & \textbf{5.7}            & \textbf{2.1}               & \textbf{12.2}        & \textbf{393ms}       \\ \hline
Ground-truth (Switchboard)                & 4.3             & 1.3                & 4.4          & 518ms       \\ \hline
\end{tabular}
\caption{Naturalness evaluation results in a multi-turn setting}
\label{tab:s2s_ipus}
\end{table*}

\begin{table*}[t]
\centering
\begin{tabular}{lcccccccc}
\hline
         & Training set                                                           & \multicolumn{5}{c}{Next action prediction}                                                                                                                                                                                                                                  & \multicolumn{2}{c}{Response generation}                                                                 \\ \hline
         &                                                                        & SIL                                                 & SPK                                                 & CON                                                 & STP                                                 & ACC                                                 & Coherence                                         & Sensible                                            \\ \hline
Cascaded & \begin{tabular}[c]{@{}c@{}}Switchboard\\    \ + Pretraining\end{tabular} & \begin{tabular}[c]{@{}c@{}}-\\ -\end{tabular}       & \begin{tabular}[c]{@{}c@{}}-\\ -\end{tabular}       & \begin{tabular}[c]{@{}c@{}}-\\ -\end{tabular}       & \begin{tabular}[c]{@{}c@{}}-\\ -\end{tabular}       & \begin{tabular}[c]{@{}c@{}}-\\ -\end{tabular}       & \begin{tabular}[c]{@{}c@{}}5.8\\ 6.6\end{tabular} & \begin{tabular}[c]{@{}c@{}}59\%\\ 69\%\end{tabular} \\ \hline
RTTL-DG & \begin{tabular}[c]{@{}c@{}}Switchboard\\  \  + Pretraining\end{tabular} & \begin{tabular}[c]{@{}c@{}}0.85\\ 0.85\end{tabular} & \begin{tabular}[c]{@{}c@{}}0.49\\ 0.52\end{tabular} & \begin{tabular}[c]{@{}c@{}}0.95\\ 0.95\end{tabular} & \begin{tabular}[c]{@{}c@{}}0.63\\ 0.63\end{tabular} & \begin{tabular}[c]{@{}c@{}}0.83\\ 0.85\end{tabular} & \begin{tabular}[c]{@{}c@{}}4.8\\ 5.2\end{tabular} & \begin{tabular}[c]{@{}c@{}}39\%\\ 45\%\end{tabular} \\ \hline
\end{tabular}
\caption{Effectiveness of pretraining on next action prediction and response generation task. SIL, SPK, CON, STP, and ACC denote Remain Silent, Initiate Speaking, Keep Speaking, Stop Speaking, and Overall Accuracy, respectively}
\label{tab:s2s_pretrain}
\end{table*}

In multi-turn evaluations, RTTL-DG also demonstrates superior performance compared to the cascaded system across all metrics, closely aligning with results observed on real data. The cascaded model, by design, produces alternating speech turns without overlaps, back-channels, or pauses, resulting in rigid, mechanical interactions. Furthermore, the cascaded system exhibits a significant average gap or latency of 800ms between turns. This delay can cause substantial user frustration and make the interaction feel unnatural when deployed in real-world scenarios. In contrast, the RTTL-DG model excels in generating overlaps, pauses, and back-channels—often at a frequency even higher than that of real conversations. The average gap duration in RTTL-DG-generated dialogues is notably lower at 393ms, compared to 518ms in real data. This suggests that responses are generated quickly and fluidly. This efficiency arises because RTTL-DG can begin formulating responses before the other speaker has finished speaking. Additionally, RTTL-DG-generated gaps are more stable and exhibit lower variance compared to those in real conversations, where gaps can range from extremely short to several seconds, skewing the average duration upward.

\subsection{Effectiveness of pretraining on synthetic data}

We evaluated how pre-training on synthetic data impacts the overall performance for both the Cascaded and RTTL-DG models. As shown in Table \ref{tab:s2s_pretrain}, pre-training with synthetic data significantly improved the response generation task for the RTTL-DG model, increasing the coherence score from 4.8 to 5.2. Notably, this improvement in response generation also enhanced the next action prediction task, with the accuracy score rising from 83\% to 85\%. This suggests that understanding the conversation's semantics plays a crucial role in turn-taking decisions.

Similarly, a more substantial improvement was observed in the Cascaded model. This can be attributed to the fact that the Cascaded model uses text as the target response, which is the same for both the pre-training and fine-tuning stages. In contrast, the RTTL-DG model uses speech units as the target response, leading to some discrepancies between pre-training and fine-tuning. This is because the speech units in synthetic speech might not consistently align with the real speech units, as mentioned earlier.

Overall, pre-training on synthetic data resulted in significant improvements for the RTTL-DG model, suggesting that expanding the training data or designing more effective synthetic data generation and augmentation techniques could lead to further advancements.

\section{Related works}
\label{sec:related_work_responsive}

Current spoken dialogue systems, while advanced in many aspects, still fall short of achieving the naturalness characteristic of human conversation. Their responses tend to lack the fluidity and spontaneity of human speech, frequently sounding stilted and overly formal. As most spoken systems are built on a cascaded design consisting of three separate components—ASR, DRG, and TTS—they have difficulty managing natural turn-taking, such as handling interruptions and interpreting non-verbal cues, which disrupts the flow of interactions. Several approaches have been proposed to enhance conversational fluidity, and effective turn-taking to improve the naturalness and user experience.

\subsection{Turn-taking modeling}

Turn-taking is a fundamental aspect of human conversation, enabling seamless and intuitive interactions. Nothdurft et al. \cite{nothdurft2014finding} emphasized the challenges involved in modeling turn-taking, exploring questions such as whether it is necessary, when it should be incorporated into ongoing interactions, and how it should be implemented. Traditional dialogue systems often rely on simplistic mechanisms such as fixed silence thresholds to determine turn shifts, which can result in awkward pauses, interruptions, or overly long responses that disrupt the conversational flow. Turn-taking modeling refers to the process of predicting and managing when each participant in a conversation should speak or respond. Instead of relying solely on silence indicators, modern approaches incorporate a variety of cues to enhance turn-taking accuracy. Lexico-syntactic cues involve analyzing the structure and choice of words to predict when a speaker is about to finish their turn \cite{lexical_cues}. Schlangen et al. \cite{schlangen2006reaction} identify word-final pitch and intensity levels, along with $N$-grams-based features, as potent predictors for turn-taking. Building on this foundation, Atterer et al. \cite{atterer2008towards} enhance the accuracy of classifying words as utterance-final or non-final. They highlight features, specifically word and part-of-speech n-grams, as the most important features. Prosody, namely, the rhythm and intonation of speech, provides crucial information about the speaker's emotional state and intent, helping the system to determine natural points for turn transitions. Erik et al. \cite{ekstedt2022much} train voice activity projection models using an extensive dataset of two-person conversations to capture prosodic features relevant to turn-taking. They also conduct various analyses to show when and how prosodic information is crucial for predicting turn-taking. Gaze direction and other non-verbal cues, such as facial expressions and gestures \cite{gaze_cues}, also play a significant role in human communication and are increasingly being used to inform turn-taking decisions in dialogue systems.

\subsection{Back-channel generation}

In human conversations, back-channel responses are crucial because they provide real-time feedback to the speaker, indicating that the listener is attentive, understanding, and emotionally engaged. Examples of back-channel cues include short phrases like "mm-hmm," "yeah," and other non-verbal sounds of acknowledgment. Without these signals, conversations can feel one-sided and mechanical, significantly diminishing the overall user experience.

Backchannel (BC) prediction is the task of identifying appropriate moments for a listener to provide feedback and determining the type of feedback to be given. This involves two tasks: Backchannel Opportunity Prediction (BOP) and Backchannel Category Prediction (BCP). BOP identifies when a backchannel response is appropriate, while BCP classifies the type of response. Approaches have evolved from rule-based systems to advanced neural networks. Early influential work by Ward et al. \cite{ward_bc_acoustic} and Fujie et al. \cite{fujie_bc_acoustic} relied on acoustic features like pitch and pause length, with Mel Frequency Cepstral Coefficients (MFCCs) becoming standard measures. The integration of lexical features, initially through simple embeddings like Word2Vec, led to state-of-the-art performance \cite{ortega_bc_embedding}. Combining acoustic and lexical features, especially with pre-trained models like BERT, has further improved accuracy \cite{jang_bc_bert}. Multi-task learning approaches \cite{hara_bc_multi, ishii_bc_multi}, which combine BC prediction with related tasks like emotion classification and turn-taking prediction, have highlighted the benefits of leveraging the multifaceted nature of dialogue for more accurate BC prediction.

\subsection{Duplex conversation modeling}
Duplex conversation modeling integrates turn-taking and back-channel generation, which aims to make the conversational experience as seamless as possible, allowing people to speak as they would with another person. For instance, Google Duplex integrates back-channel responses like "um-hum" and "well..." into its dialogue system for booking restaurants, which greatly improves the natural flow of conversations. Jin et al. \cite{jin_duplex_outbound} present an outbound agent that can detect user interruptions and fragmented speech, while Microsoft Xiaoice \cite{xiaoice} features a rhythm control module to enhance turn-taking in intelligent assistants. Similarly, Alibaba's Duplex Conversation system \cite{alibaba_duplex}, a multi-turn, multimodal spoken dialogue system, facilitates smooth telephone-based interactions through three key subtasks: user state detection, back-channel selection, and barge-in detection.

\subsection{End-to-end duplex dialogue systems}

A cascaded duplex conversation model comprising multiple modules—such as ASR, DG, TTS, turn-taking prediction, and BC generation—can encounter several weaknesses. These include increased latency from processing delays in each module, error propagation where mistakes in one component affect the entire system, and complexity in coordinating between modules. Additionally, since each module is typically trained independently using different datasets, there can be significant inconsistencies and discrepancies in distribution when integrating them together, especially when the output of one model serves as the input for another module. Finally, the model can be resource-intensive and inflexible, demanding substantial computational power, memory, and storage, which can lead to overhead in system development and maintenance.

Recent approaches have focused on integrating all modules into a single, unified model, trained in an end-to-end fashion whereby all tasks are optimized jointly. In task-oriented dialogue systems, end-to-end spoken language understanding systems predict utterance semantics directly from speech using a single model, eliminating the need for a separate speech recognition model. The Speech and Language Model (SLM) integrates a pretrained speech model with a LLM into a unified framework, enabling direct processing of audio inputs and language outputs. A retrieval-augmented SLM builds on the SLM by incorporating a retrieval mechanism to address the challenge of recognizing domain-specific entities, allowing dialogue states to be inferred directly from the audio signal. These end-to-end models reduce latency, minimize error propagation, and simplify system coordination. In open-domain dialogue systems, Tu et al. introduced dGSLM \cite{dgslm}, the first "textless" model capable of generating naturalistic spoken dialogue from a conversation prompt between two speakers. The model leverages recent advancements in unsupervised spoken unit discovery \cite{gslm}. It employs a dual-tower transformer architecture and is trained on 2,000 hours of two-channel raw conversational audio from the Fisher dataset \cite{fisher}, without any accompanying text or labels. The trained model generates speech, laughter, and other paralinguistic signals in both channels simultaneously, producing more natural and fluid turn-taking compared to text-based cascaded models. However, the semantic coherence of the generated dialogue remains low, and the model can only produce responses offline. This limitation makes it challenging to deploy in real-world chatbot applications that require real-time response generation based on the user's audio input. 

To address these challenges, many recent methods \cite{defossez2024moshi,xie2024mini, wang2024full, zeng2024glm, zhang2024omniflatten, meng2024parrot, chen2024slam} have been proposed, leveraging powerful pretrained LLMs to enhance the semantic quality of responses. These approaches also integrate speech or acoustic tokens to improve the naturalness and expressiveness of generated speech. Additionally, these next-generation models demonstrate full-duplex capabilities, enabling seamless turn-taking and significantly lower latency compared to traditional cascaded models.

\section{Conclusion}

This paper focuses on improving the responsiveness and expressiveness of spoken dialogue generation. Unlike traditional cascaded dialogue systems, which generate conversations in a turn-by-turn manner, we introduce a real-time textless dialogue generation model. This model is designed to produce faster, more natural responses by incorporating conversational elements such as pauses, hesitations, backchannels, and overlaps. Experimental results show that while the RTTL-DG model underperforms the cascaded model in terms of semantic coherence, it significantly better in  naturalness aspect. Specifically, the dialogues generated by the RTTL-DG model closely mimic real human conversations. We believe this represents a vital step toward the creation of more human-like dialogue systems, with potential applications that could greatly enhance user experiences in conversational AI.

\section*{Acknowledgements}
This publication has emanated from research supported in part by a grant from Science Foundation Ireland under Grant number 18/CRT/6183. For the purpose of Open Access, the author has applied a CC BY public copyright licence to any Author Accepted Manuscript version arising from this submission.

\bibliography{anthology,custom}
\bibliographystyle{acl_natbib}

\end{document}